\documentclass[10pt,twocolumn,letterpaper]{article}

\usepackage{iccv_arxiv}
\usepackage{times}
\usepackage{epsfig}
\usepackage{graphicx}
\usepackage{amsmath}
\usepackage{amssymb}
\usepackage{bm}

\usepackage{caption}
\usepackage{booktabs}
\usepackage{multirow}
\usepackage{xcolor}
\usepackage{pifont}

\definecolor{m_green}{HTML}{2C8915}
\definecolor{m_red}{HTML}{FF0000}
\newcommand{\cxmark}{\textcolor{m_red}{\ding{55}}}
\newcommand{\checkmarkgreen}{\textcolor{m_green}{\ding{51}}}

\usepackage[pagebackref=true,breaklinks=true,letterpaper=true,colorlinks,bookmarks=false]{hyperref}

\iccvfinalcopy 

\def\iccvPaperID{28} 

\ificcvfinal\fi

\begin{document}

\title{Self-supervised Hypergraphs for Learning Multiple World Interpretations}

\author{Alina Marcu\textsuperscript{1,2} \hspace{1cm}
Mihai Pirvu\textsuperscript{1,2} \hspace{1cm}
Dragos Costea\textsuperscript{1,2} \hspace{1cm}
Emanuela Haller\textsuperscript{3} \hspace{1cm}
\vspace{1mm}\\
Emil Slusanschi\textsuperscript{1} \hspace{1cm}
Ahmed Nabil Belbachir\textsuperscript{4} \hspace{1cm}
Rahul Sukthankar\textsuperscript{5} \hspace{1cm}
Marius Leordeanu\textsuperscript{1,2,4}
\thanks{Primary contact: Marius Leordeanu at leordeanu@gmail.com}
\vspace{1.5mm}\\
\textsuperscript{\rm 1}UPB \hspace{1cm}
\textsuperscript{\rm 2}IMAR \hspace{1cm}
\textsuperscript{\rm 3}Bitdefender \hspace{1cm}
\textsuperscript{\rm 4}NORCE \hspace{1cm}
\textsuperscript{\rm 5}Google Research \hspace{1cm}
}

\maketitle
\ificcvfinal\thispagestyle{empty}\fi


\begin{abstract}
We present a method for learning multiple scene representations given a small labeled set, by exploiting the relationships between such representations in the form of a multi-task hypergraph. We also show how we can use the hypergraph to improve a powerful pretrained VisTransformer model without any additional labeled data. In our hypergraph, each node is an interpretation layer (e.g., depth or segmentation) of the scene.
Within each hyperedge, one or several input nodes predict the layer at the output node. Thus, each node could be an input node in some hyperedges and an output node in others. In this way, multiple paths can reach the same node, to form ensembles from which we obtain robust pseudolabels, which allow self-supervised learning in the hypergraph. We test different ensemble models and different types of hyperedges and show superior performance to other multi-task graph models in the field. We also introduce Dronescapes, a large video dataset captured with UAVs in different complex real-world scenes, with multiple representations, suitable for multi-task learning.
\end{abstract}
\vspace{-4mm}
\section{Introduction}

Learning robustly multiple interpretations of the complex world, such as segmentation, depth, and surface information, with minimal human supervision is one of the great challenges in vision today. 
In this work, we exploit the consensus that naturally appears between such interpretation layers in order to learn them self-supervised. We construct a multi-task hypergraph, in which nodes represent the layers, while the hyperedges (or edges) group them together and capture their relationships. Each hyperedge has one or multiple input nodes that are transformed, through a neural network, into a single output one. Thus, the set of hyperedges forms multiple pathways through the hypergraph that could reach a given node. This gives the possibility to
form ensembles from which robust pseudolabels can be extracted for the output node when supervised data is not available. Such pseudolabels, along with the available ground truth, are then used as a supervisory signal to distill self-supervised single hyperedges for the next learning cycle. With each cycle, we add novel unlabelled data in order to follow a scenario which is often met in practice, when more data is easily available but annotations are not.

In our extensive experiments, we demonstrate the effectiveness of our contributions: we show that more complex hyperedges bring a strong boost over simple pairwise edges. We also show that by adding an unsupervised learning cycle on unlabeled data from new scenes, we improve generalization and performance on those scenes. Moreover, we demonstrate that the multi-task consensus, which improves during the unsupervised learning stages, also brings the desired temporal stability and consistency, even though no explicit temporal information is used in our framework and all processing is done at the level of single frames. Last but not least, we show that our self-supervised hypergraph learning can be used to improve in both accuracy and temporal consistency over a heavily trained transformer expert when such an expert is used to initialize our hyperedge (edge) nets. This result is even more surprising when we consider the fact that our neural nets are lightweight U-Nets with two orders of magnitude fewer parameters.

Our \textbf{main contributions} are: 
\vspace{-1.5mm}
\begin{itemize}
  \item We introduce, to the best of our knowledge, the first multi-interpretation hypergraph model that considers higher-order relationships between multiple world views and learns to find consensual pseudolabels from multiple pathways in the hypergraph. In extensive experiments, we show that our model is superior to related multi-task self-supervised graph models, which consider only simple edges and do not learn the ensembles that form pseudolabels.
\vspace{-2mm}
  \item We introduce Dronescapes, a large video dataset with annotations for semantic segmentation, odometry, and 3D information. This represents a very complex real-world test bed, with a wider variety of scenes, which distinguishes our work with respect to the self-supervised multi-task literature, which is limited to synthetic or simpler real-world contexts (e.g., indoors).
\vspace{-4mm}
  \item We demonstrate a solid generalization capability, by learning multiple tasks, with human supervision available only for the segmentation task for approximately 1$\%$ of all training data. Moreover, our hypergraph uses lightweight neural nets which can significantly improve both the accuracy and the temporal consistency over powerful, heavily pretrained experts, even though no temporal information is used.
  \vspace{-4mm}
\end{itemize}

\begin{figure}
\centering
\includegraphics[scale=0.25]{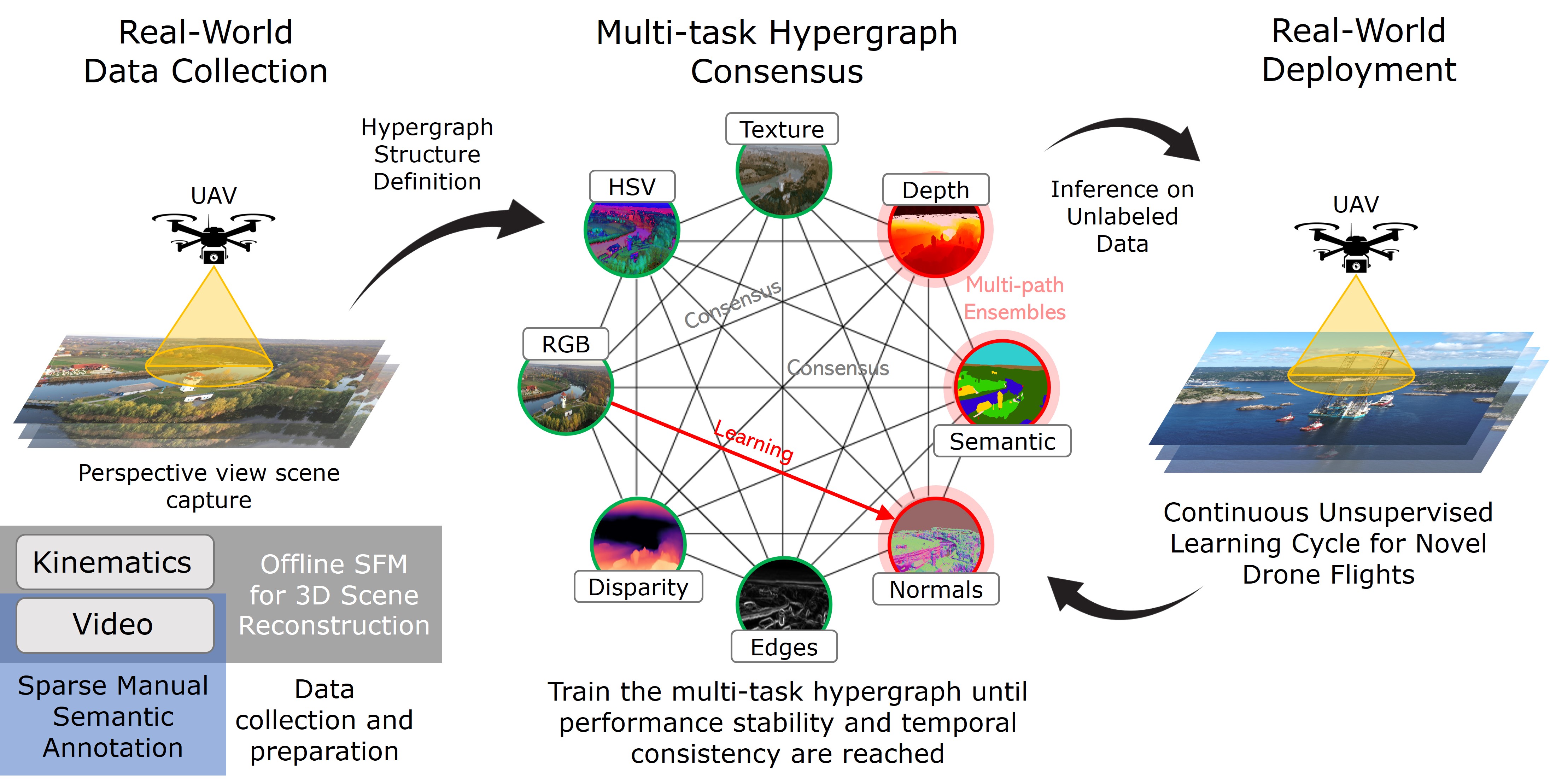}
\caption{Our self-supervised multi-layer hypergraph in the context of real-world UAVs. We use kinematics and video drone data for offline 3D scene reconstruction and sparse manual labeling as an initial data collection and annotation for three dense prediction tasks: semantic segmentation, depth and surface normals estimation. We define the hypergraph structure in terms of input nodes (from sensors) and output nodes (those that will be predicted). We train in a self-supervised manner over multiple cycles,
which improves both accuracy and temporal consistency. The hypergraph is also able to adapt to novel scenes and improve over initial powerful experts used for initialization.}
\label{fig:main_figure}
\vspace{-4mm}
\end{figure}

\section{Related work}\label{sec:related_work}

\noindent\textbf{Consensus-based learning:} Recent works~\cite{regatti2021consensus,grill2020bootstrap} exploit clustering-based consensus from different transformations of the same image, but our goal is to consider meaningful scene representations (tasks), not just representation clusters. Also, focusing on a single task degrades the performance of others. It is difficult to achieve all-around improvements, as we aim here. In order to better balance the tasks, some approaches assign them weights~\cite{liu2022auto}. Other works ~\cite{tosi2020distilled} take a step further towards holistic scene understanding by learning together, and self-supervised, depth and motion alongside semantics. Our model is more general and can accommodate in principle any number of tasks.

\noindent\textbf{Graph-based learning:} Work on hypergraph neural networks is scarce and uses a single label for a node~\cite{feng2019hypergraph, tudisco2021nonlinear}. Iterative graph-based semantic segmentation is also related, but it considers only one or two representations~\cite{marcu2020semantics,zhen2020joint}. Other semi-supervised multi-task approaches do not consider higher-order relations between tasks and do not learn the ensembles used for generating pseudolabels~\cite {leordeanu2021semi,haller2021cshift}. Other works do not have a semi-supervised learning component, but focus on robustness instead, with two-hop graphs~\cite{Yeo_2021_ICCV} or larger multi-task ones~\cite{zamir2020robust}. Other related work~\cite{zamir2018taskonomy} shows there is a strong relationship between multiple complementary tasks and that simultaneously exploiting the common and distinctive features between these tasks is effective for generalizing to out-of-distribution scenes.

\noindent\textbf{Unsupervised multi-task learning:} There are several important advances in unsupervised learning that are based on constraining together several specific tasks such as relative pose, depth and even semantic segmentation~\cite{chen2019self, zhou2017unsupervised, ranjan2019competitive, bian2019unsupervised, gordon2019depth, yang2018unsupervised, tosi2020distilled, guizilini2020semantically, chen2019towards, stekovic2020casting}. There are also approaches that combine inputs from multiple senses for cross-modal prediction learning~\cite{Hu_2019_CVPR, Li_2019_CVPR, zhang2017split, pan2004automatic, he2017unsupervised, zhao2018sound}. 

\noindent\textbf{Knowledge Distillation:} The Teacher-Student paradigm is one of the most popular approaches in which smaller networks have been shown to be very effective in learning from large networks~\cite{liu2019structured}. However, it is very rare that the smaller Student outperforms the Teacher~\cite{wang2021knowledge}. A few methods perform knowledge distillation from multiple representations but without the graph structure~\cite{cai2021x}. Feeding pseudolabels for retraining the Student or Teacher is known as self-distillation~\cite{gou2021knowledge}. Some methods perform distillation to yield compact models suitable for real-time inference~\cite{wu2021real}. Others attempt to use network architecture search (NAS) to design better models that achieve the same objective~\cite{chen2019fasterseg}. Nevertheless, the best-performing models are trained from scratch, either by developing a lighter version of top-performing architectures~\cite{zhang2022topformer} or by optimizing older architectures suitable for parallel operations~\cite{gao2022fbsnet}.

\noindent\textbf{Our work in context:} we put together and advance over many previous ideas, with demonstrated benefits. We do self-distillation, as we train single hyperedges at the next iteration by ensemble teachers from the previous one. We propose a hypergraph structure, as the only way to consider many tasks and capture their complex relationships, and also form supervisory ensembles for each, within a single system. We do self-supervised learning, by letting the graph teach itself, through such unsupervised ensembles, after an initial stage of learning either from experts or strictly supervised with limited labeled data.

\section{Multi-task Self-supervised Hypergraphs}\label{sec:multitask_section}

We present an overview of the proposed hypergraph model 
in Fig.~\ref{fig:main_figure}. Hypergraph nodes represent different interpretation layers of the scene (e.g. RGB, semantic segmentation, depth, camera normals, etc.). Hyperedges capture relations between two or more nodes. Thus, in a k-th order hyperedge, a neural net takes k-1 node layers as input and learns to output the k-th node layer. In this manner we obtain, in the same hypergraph,
multiple paths (by going through intermediate hyperedges) from the sensor input nodes (e.g. RGB images) to any output node. In principle, any node could play simultaneously the role of an output or an input, in different hyperedges. Thus, for a given output node, we have several candidate layers, coming from different paths, which are used to form an ensemble. We use these ensembles, at the current learning cycle, to produce pseudolabels to distill (retrain) the edges and hyperedges in the next iteration. In experiments, we observed that when more unlabeled data is added, each self-supervised learning cycle improves not only accuracy but also temporal consistency, which, as discussed in Section~\ref{sec:experiments}, could be interpreted as a measure of trustworthiness.

\subsection{Structure of Hyperedges}
\label{sec:hyperedge_types}

\begin{figure}[!t]
\centering
\includegraphics[scale=0.38]{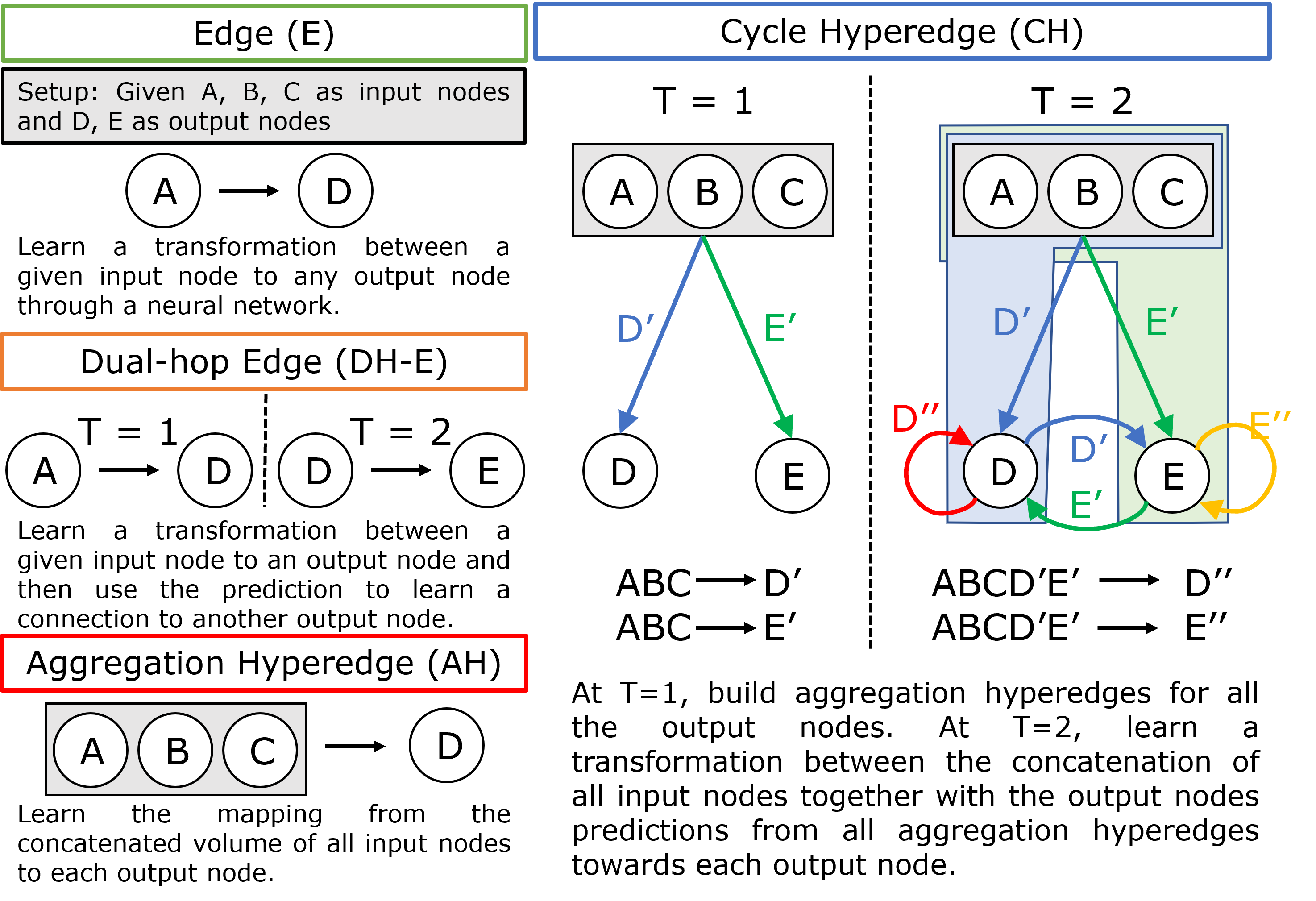}\vspace{-2mm}
\caption{Types of edges and hyperedges in the hypergraph. Previous works use only edges, while we introduce two types of hyperedges to capture more complex relations between different layers.}
\label{fig:hyperedges_types}
\vspace{-4mm}
\end{figure}

In our hypergraph, each hyperedge (often referred to as ``edge`` in the 2nd order case) has a set of one or multiple input nodes and a single output one. Input nodes could be either known (from sensors) or pseudolabels estimated by ensembles as explained in Sec. \ref{sec:learning_ensembles}.
Hyperedges are modeled by lightweight U-Nets~\cite{ronneberger2015u}, with the same structure of ~1.1M parameters, but independently learned. Below we present the different types of hyperedges that we introduce (also see Fig.~\ref{fig:hyperedges_types}): 
\vspace{-2mm}
\begin{itemize}
    \item \textbf{Edges (E)} - they learn a transformation between an input node to an output node through a neural net, similar to recent work on self-supervised multi-task graph models~\cite{leordeanu2021semi, haller2021cshift, zamir2020robust, zamir2018taskonomy}. We name each edge based on its input and output node layers (e.g., \textit{rgb $\rightarrow$ sseg} refers to the edges that have the RGB image as input and the segmentation layer as output). Such direct edges from RGB input are the main ones we aim to improve during hypergraph learning. They are light and can be deployed at a small cost in practice (e.g., on UAVs).
    \vspace{-2mm}
    \item \textbf{Dual-hop Edges (DH-E)} - they use intermediate predicted representations. For example, the dual-hop edge \textit{rgb$\rightarrow$depth$\rightarrow$sseg} refers to the path that takes RGB as input, it produces the depth layer, and from that depth layer, it outputs the segmentation layer. 
    \vspace{-2mm}
    \item \textbf{Aggregation Hyperedges (AH)} - they concatenate all input representations and learn the mapping from the aggregated volume of input nodes to each output node. We denote AH-ufo as the hyperedge that includes at input the prediction of UFODepth, an off-the-shelf purely unsupervised depth estimation method~\cite{licuaret2022ufo}, to better understand the impact of such methods, concatenated with the RGB image.
    \vspace{-2mm}
    \item \textbf{Cycle Hyperedges (CH)} - they start from the concatenation of all input nodes together with the outputs of all AH hyperedges (including the AH output for the current node) and predict the output at the current node. 
\vspace{-3mm}
\end{itemize}

\subsection{Learning Hyperedge Ensembles}\label{sec:learning_ensembles}

As data passes from all input nodes ($N_i$) to all output nodes $(N_o)$ through the hypergraph $G(N_i, N_o)$ via the edge/hyperedge neural nets, various paths are created. At the end of the message passing process (\emph{MP}), each output node has a list of messages, one for each such path. These messages can be seen as candidates in an ensemble learning process to compute a final response. Traditionally, ensembles are formed by simply averaging the candidates, in the regression case, or taking the majority, in the classification case. We study whether this ensemble mechanism can be improved by adding a secondary learning process, one at each output node, on top of the pathway-independent candidates. This can be seen as a special case of the~\textit{aggregate} function in Graph Neural Networks (GNNs): $\mathbf{Y}_{agg}(n_o) = f_{agg}(\mathbf{Y}_{n_o})$, where $\mathbf{Y}_{n_o} = \mathit{MP}(n_i \xrightarrow{} ... \xrightarrow{} n_o)$, for each pathway defined in the graph structure, with data $\mathbf{X}_{N_i}$ provided only at input nodes.

We introduce 2 types of parameterized ensemble models: \textit{Linear Ensembles} and \textit{Neural Network Ensembles}. Note that all individual edges are pretrained and frozen. The second optimization is done only on the results of each pathway, on the labeled training set, for which we have access to ground truth (denoted with $\mathbf{gt}$). The tensor shape of each output node before aggregation is $\mathbf{Y_{n_o}} :: (p_{n_o}, n, c_{n_o}, h, w)$, where $p_{n_o}$ represents the number of pathways, $n$ is the train set size and $(c_{n_o}, h, w)$ represents the shape of each output representation.

\noindent{\textbf{Linear Ensembles:}} We want to learn a vector of weights $\mathbf{w}$, with one $w_i$ per pathway $i$, using the train set, which is going to be used later on, to produce pseudolabels on the semi-supervised set. The weights are learned using linear regression. For the semantic segmentation node, we optimize using logistic regression.
The aggregation function becomes the weighted sum: $\mathbf{Y}_{agg} = \mathbf{w^\top Y_{n_o}}$. We repeat this process for each output node, resulting in a fixed weights matrix $\mathbf{W}$ which is then used for generating pseudolabels.

\begin{figure}
  \centering
  \includegraphics[width=0.9\linewidth]{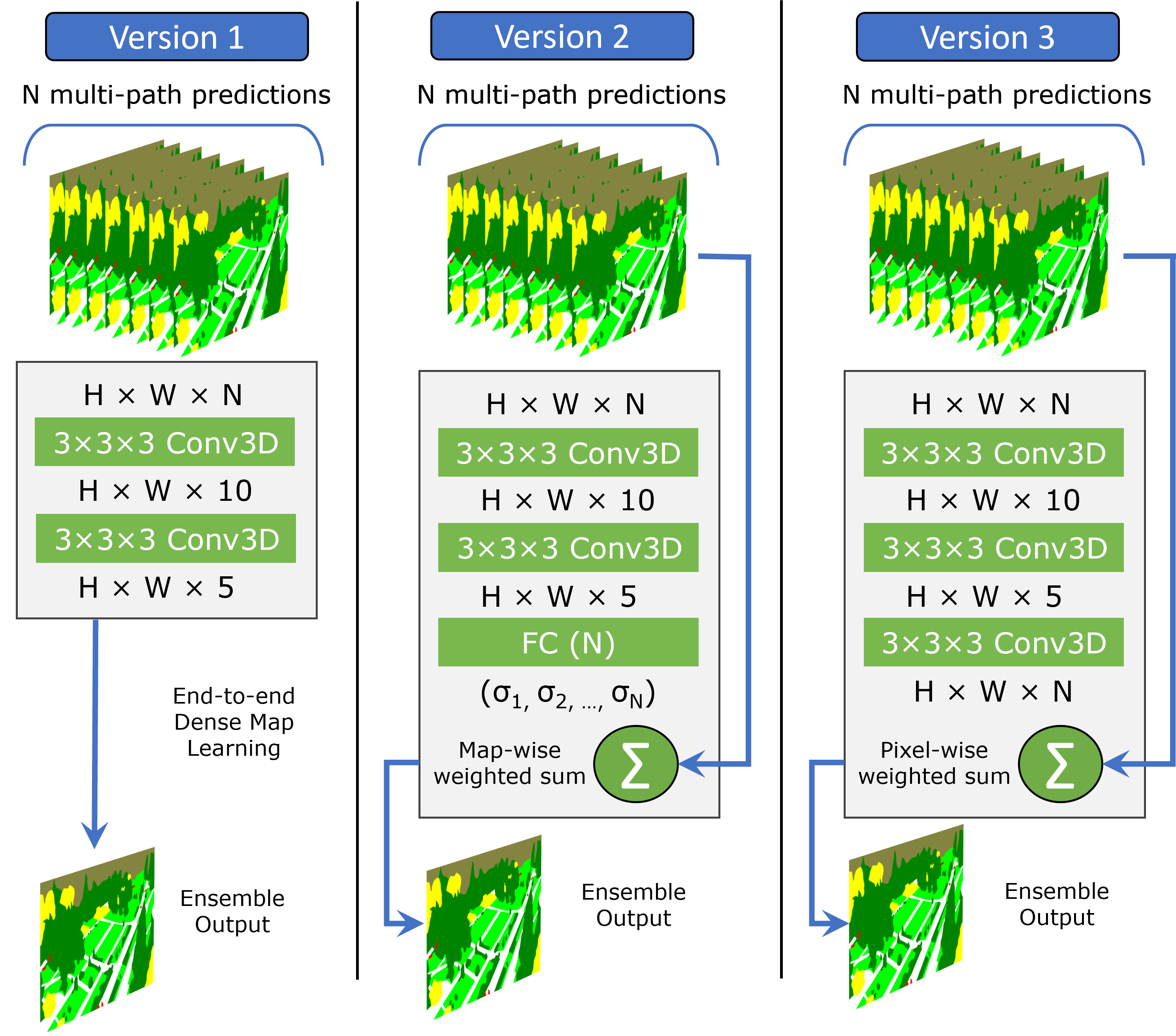}
  \vspace{-2mm}
  \caption{Proposed Neural Network Ensembles: \textbf{Version 1} produces directly a final output map. \textbf{Version 2} outputs one weight per each candidate input, then adds the weighted inputs. \textbf{Version 3} outputs one weight per each pixel for each candidate input, then adds the weighted inputs.}
\label{fig:conv3ds-neural-aggregator}
\vspace{-4mm}
\end{figure}

\noindent{\textbf{Neural Network Ensembles:}}  Instead of learning a set of fixed weights, as before, in this case we train a separate neural net (\emph{NN}) for each output node. Thus, each new input will dynamically produce, through this neural net, a new way of combining the set of candidate outputs $\mathbf{Y}$ (defined as before). The loss function is $\min|(\mathit{NN}(\mathbf{Y}) - \mathbf{gt})|$ and the \emph{NN} model is optimized using iterative gradient descent.
We propose three types of neural net ensembles (as also shown in Fig.~\ref{fig:conv3ds-neural-aggregator}):
\textbf{Version 1).} \emph{NN} produces directly a final output map: $Y_{agg} = \mathit{NN}(\mathbf{Y})$. \textbf{Version 2).} Dynamic weights, one for each
candidate layer, aggregated by weighted mean: $\mathbf{y_{NN}} :: (p,) = \mathit{NN}(\mathbf{Y})$; $Y_{agg} = \mathbf{y_{NN}}^T \mathbf{Y}$. \textbf{Version 3).} Dynamic weights, a separate one for each pixel in each candidate layer, followed by point-wise multiplication: $\mathbf{Y_{NN}} :: (p, c, h, w) = \mathit{NN}(\mathbf{Y})$; $Y_{agg} = \mathbf{Y_{NN}} \odot \mathbf{Y}$.

\subsection{Self-supervised iterative hypergraph learning}\label{subsec:semisup_hypergraph_learning}

Learning in the hypergraph requires an initial set of annotations, which is either given by humans, automatically generated (by an offline analytical method) or provided by a pretrained expert. These annotations are used to initialize the individual edges and hyperedges and then learn the ensemble functions. Once the initialization stage is completed, we can proceed with the semi-supervised learning stages (iterations), in which we first produce pseudolabels on newly added data, then retrain the individual edges and hyperedges on the new pseudolabels (including all the other labels and pseudolabels available from previous iterations). Succinctly, each self-supervised learning iteration consists of 1) adding new unlabeled data; 2) producing pseudolabels for the new data; 3) retraining the hyperedges by including the new pseudolabeled data in the training set. 


\subsection{Dronescapes Dataset}\label{sec:dronescapes_dataset}

\begin{figure*}
\centering
\includegraphics[scale=0.073]{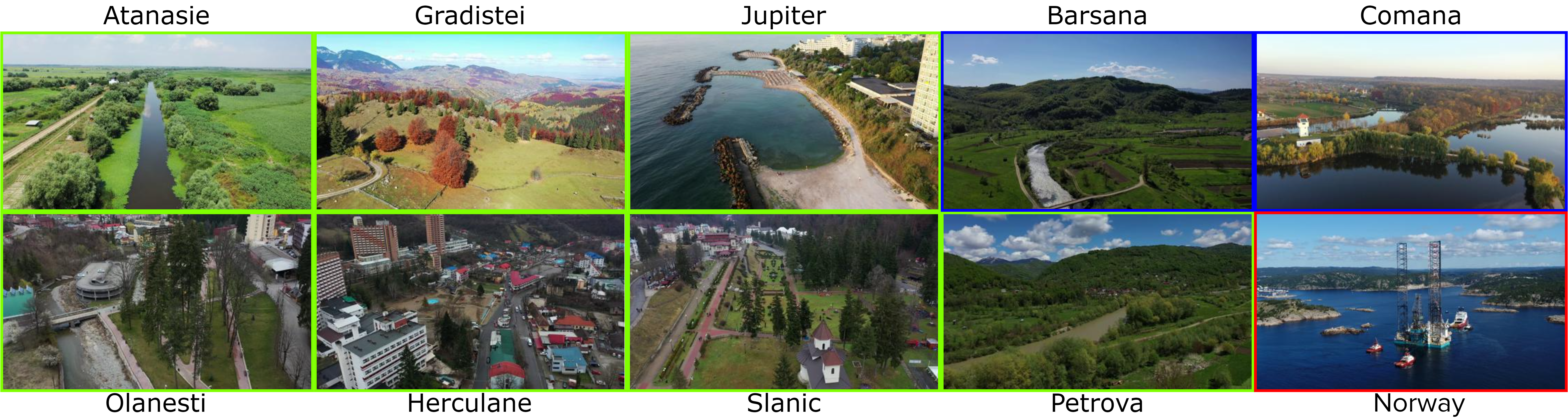}
\caption{Sample frames from each of the 10 scenes from the Dronescapes dataset. The scenes framed with \textbf{\color{green} green} borders represent training scenes for which we have access to a small fraction of manual annotations during training. The others depict unseen, test scenes with semantic distributions that are closer to the training set (in \textbf{\color{blue} blue}) or out-of-distribution (\textbf{\color{red} red}). There is a large variation in spatial distributions of classes among the different Dronescapes scenes, which range from rural (Atanasie, Gradistei, Petrova, Barsana, Comana), to urban (Olanesti, Herculane, Slanic) and seaside (Jupiter, Norway), while also being geographically far apart.} 
\label{fig:samples_dataset}
\vspace{-4mm}
\end{figure*}

We introduce a large-scale UAV video dataset with automatic odometry and 3D information for all frames and
semi-automatic semantic segmentation annotation for a subset of frames, as explained later in this Section. 
All video sequences include GPS information, linear and angular velocities, and absolute camera angles. The total length is about $50$ minutes. Videos have $3840\times2160$ $30$ FPS images, while the odometry is provided at 10 Hz. We collect a total of 10 widely varied scenes that we split into $7$ training and $3$ test scenes. A visual representation for each scene from our dataset is shown in Figure~\ref{fig:samples_dataset}. We highlight the variety in landscapes, altitudes, and object scales between each of the rural (Atanasie, Gradistei, Petrova, Barsana, Comana), urban (Olanesti, Herculane, Slanic) and seaside scenes (Jupiter, Norway). For the test scenes we chose three different kinds: one (Barsana) with a more similar semantic class distribution to at least one of the training scenes, one (Comana) that is less similar to the training scenes and a third one (Norway), that is very different from any of the training scenes.


\noindent \textbf{Dataset split.} Exclusively for the task of semantic segmentation, we sparsely manually annotate frames. Based on their number we divide the scenes in \textit{weakly-labeled scenes} (Gradistei, Herculane, Jupiter, Petrova, Olanesti, Barsana and Comana) and \textit{strongly-labeled scenes} (Atanasie, Slanic and Norway). For the strongly labeled scenes, we sample frames every $2$ seconds covering the whole video. For the weakly-labeled scenes, we uniformly sample triplets of frames from each scene, such that the frames in each triplet are $2$ seconds apart (or $60$ frames between the triplet limits), while the triplets are at least $26$ seconds apart and have significant changes in pose (viewpoint). 
We divided our dataset into $4$ sets, in a suitable manner for multiple training iterations with the addition of novel data every iteration. In Tab.~\ref{tab:dataset_split} we present these splits and the total number of frames per scene, alongside the number of manually labeled frames. Half of the triplets/frames from the training scenes are included in \textit{Train Unlabeled (iter 1)}, whilst the other half is in \textit{Train Unlabeled (iter 2)}. The frames from all the test scenes are included in \textit{Train Unlabeled (iter 3)}. We note that the only manually-labeled frames used in learning the hypergraph are the frames from \textit{Train Labeled} set, whilst the rest are used for evaluation purposes only. 


\begingroup
    \setlength{\tabcolsep}{2pt} 
    \begin{table}
        \centering
        \caption{Dronescapes dataset split. For each training scene we report the number of frames used for purely supervised training (Train Labeled set), and also for iterative self-supervised training (Train Unlabeled iterations 1, 2 and 3 sets). In parenthesis we show the number of frames for which we have manual segmentation annotations. Except for the Train Labeled set on the training scenes, all the other annotations are used strictly for evaluation.}
        \label{tab:dataset_split}
        \begin{tabular}{lcccc}
            \toprule
            \shortstack{Scene\\Name} & \shortstack{Train\\Labeled}  & \shortstack{Train\\Unlabeled\\(iter 1)} & \shortstack{Train\\Unlabeled\\(iter 2)} & \shortstack{Train\\Unlabeled\\(iter 3)}\\
            \midrule
             Atanasie & 76 & 4501\hspace{1mm}(76) & 4500\hspace{1mm}(75) & -\\
            Gradistei & 18 & 726\hspace{2mm}(18) & 484\hspace{2mm}(12) & - \\
            Herculane  & 12 & 484\hspace{2mm}(12) & 363\hspace{2mm}(9) & - \\
            Jupiter & 21 & 847\hspace{2mm}(21) & 605\hspace{2mm}(15) & - \\
            Olanesti & 18 & 726\hspace{2mm}(18) & 484\hspace{2mm}(12) & - \\
            Petrova & 12 & 484\hspace{2mm}(12) & 363\hspace{2mm}(9) & - \\
            Slanic & 76 & 4501\hspace{1mm}(76) & 4500\hspace{1mm}(75) & - \\
            \midrule
            Barsana & - & - & - & 1452\hspace{1mm}(36)\\
            Comana & - & - & - & 1210\hspace{1mm}(30) \\
            Norway & - & - & - & 2941\hspace{1mm}(50) \\
            \midrule
            TOTAL & 233 & 12269 (233) & 11299 (207) & 5603 (116) \\
            \bottomrule 
            \vspace{-5mm}
        \end{tabular}
    \end{table}
\endgroup

\noindent\textbf{Semantic segmentation annotation.} Every pixel in a frame is labeled with one of the 8 classes - \textit{land}, \textit{forest}, \textit{residential}, \textit{road}, \textit{little-objects}, \textit{water}, \textit{sky} and \textit{hill}. The annotations process was difficult, especially due to the fact that some objects are too small to be seen clearly, while other larger regions can fall into multiple categories (e.g., an area can be labeled as hill, land, and forest at the same time). 

\noindent\textbf{Depth annotation.} The drone trajectory computed with structure from motion (SfM)~\cite{Meshroom}
is aligned automatically with the trajectory from GPS, by a similarity transformation (translation, rotation and scale), which is then applied to the SfM 3D model. By combining it with the known 6D pose (from GPS and odometry) and camera intrinsic parameters, we obtain accurate metric depth maps (less than $2\%$ error in our extensive offline tests). 
The generated metric depth maps are used for training only during Iteration 1, otherwise used for evaluation purposes.

\noindent\textbf{Surface normals annotation.} We automatically processed the SfM 3D meshes in Blender and obtained surface normals at every pixel w.r.t world coordinates, which, multiplied with the inverse of the camera rotation matrix give normals w.r.t camera (aka. "camera normals").

\noindent\textbf{Hypergraph learning on Dronescapes:} We focus on scenarios where exact ground truth or human annotations are very scarce or simply not available, which is a very common case in practice.
Such an example is the case of learning from UAV videos, which is representative for our scenario and also extremely difficult due to the wide variety and complexity of scenes and camera viewpoints. As explained in the theoretical section, in each self-supervised learning cycle we add new unlabeled data. We first produce pseudolabels for each output task by using ensembles of edges and hyperedges and then retrain (distill) the next generation of edges and hyperedges by including the pseudolabeled data during training. For clarity, we detail further how we use the current dataset split in our iterative learning procedure: \textbf{Iteration 1)} Using the Train Labeled dataset, we employ a semi-automatic label propagation method~\cite{marcu2020semantics} to annotate intermediate frames from adjacent manually labeled ones and obtain Train Unlabeled (iter 1). For depth and normals, we use the 
automatically generated labels as described before. For \textbf{Iteration 2)} we used the fully-trained hypergraph from Iteration 1 and generate pseudolabels for frames in Train Unlabeled (iter 2) set, using the hypergraph ensembles for all the predicted tasks. We join the two sets Train Unlabeled (iter 1 + 2) and retrain the hypergraph. For \textbf{Iteration 3)} we repeat the steps from Iteration 2 and generate pseudolabels on the Train Unlabeled (iter 3) set, to expand the set to Train Unlabeled (iter 1 + 2 + 3).


\vspace{-2mm}

\section{Experimental Analysis} \label{sec:experiments}

We focus on learning three complementary tasks (output nodes): semantic segmentation, depth estimation and surface normals prediction. First, we study the impact made by each of our contributions w.r.t previous self-supervised graph multi-task methods, such as the addition hyperedges vs. simple edges (Sec. \ref{sec:hyperedges_vs_edges}) and 
learning parameterized multi-path ensemble models (linear and deep neural nets) vs. non-parametric ones (Sec. \ref{sec:multipath_consensus}). We also test the ability of the hypergraph to improve self-supervised over an initial state-of-the-art expert for semantic segmentation, which is used for distillation at different stages (Sec.~\ref{sec:sseg_sota_comparison}).

\noindent\textbf{Performance metrics.} We report mean IoU (\% - higher is better ($\uparrow$)) for semantic segmentation and L1 error * 100 (lower is better ($\downarrow$)) for depth and normals estimation. 

\noindent\textbf{Temporal consistency metrics.} Even though we process single frames without any temporal information, we observed that the hypergraph self-supervised learning process significantly improves the
prediction consistency between nearby frames (for pixels that belong to the same physical point), which is more than just improving average accuracy per frame. 
We designed a special \textit{consistency metric}, which uses optical flow~\cite{teed2020raft} to establish temporal chains that connect corresponding pixels across frames. For segmentation, the consistency for a given pixel measures the percentage of votes received by the winning class along the 5-frames temporal chain centered at that pixel. For depth and camera normals, we measure consistency using the variance $var$ of predictions along the same chain, as $e^{-var}$ (to map it between 0 and 1 and increase with quality). Note that if the optical flow works well, there is no occlusion (which is often the case) and the predictions are correct, then the consistencies for all three tasks should be $1$. 
We found this metric to be highly valuable, as it is complementary to the average accuracy: a predictor could 
have good accuracy on average per frame (low bias),
but lack sufficient temporal consistency (high variance). Temporally inconsistent, highly fluctuating predictions, generally indicate higher levels of uncertainty and lower reliability.

\vspace{1mm}

\noindent\textbf{Input representations.} We consider, besides the main RGB input node, other input nodes that are mathematically derived from RGB, such as HSV color, soft edges~\cite{leordeanu2014generalized} and soft segmentation~\cite{leordeanu2014generalized}), which are effective by expanding the number of edges and indirectly improving the power of the ensembles that they form at the output nodes. We also tested the possibility of adding to the pool of input nodes an unsupervised metric depth map, such as UFODepth~\cite{licuaret2022ufo}, but found it did not bring a significant boost in performance. We show examples of each of the input layer types in Fig.~\ref{fig:main_figure}.

\vspace{-1mm}

\subsection{Impact of Hyperedge Complexity}
\label{sec:hyperedges_vs_edges}

Related work tackled multi-task learning through multi-path consensus \cite{haller2021cshift, leordeanu2021semi} but did not exploit higher-order relations between tasks. We test the performance of each individual hyperedge type on all three tasks (in Tab.~\ref{tab:multitask_edges_vs_hyperedges}) and show that the higher-order ones are on average significantly stronger than the pairwise edges. With one exception: the case of metric depth, which is much more scene dependent than the other tasks and for which the more complex hyperedges overfit more easily. However, for the other two tasks, hyperedges bring a significant advantage over the simpler edges, while the dual-hop edges (DH-E), heavily used in~\cite{leordeanu2021semi}, have poor performance (since errors accumulate along the two-hops path).

\begingroup
    \setlength{\tabcolsep}{3.1pt} 
    \begin{table}
        \centering
        \caption{Evaluation of edges and hyperedges for multiple tasks: 1 - semantic segmentation (sseg); 2 - depth estimation (depth); 3 - surface normals (norm). We report mean IoU (\% - higher values are better ($\uparrow$) for the task of semantic segmentation and L1 error * 100 (lower is better ($\downarrow$)) for depth and normals estimation. The layers in the Type column, denote the input node (for edge type E) and the intermediate node (for edge type DH-E, for which RGB is always the input node). 
        We evaluate exclusively on the manually annotated frames and report mean performance over all scenes from the Train Unlabeled (iter 2) set and also on Barsana and Comana test scenes from Train Unlabeled (iter 3) (see~\ref{tab:dataset_split} for more details). Bolded results highlight the mean performance gain of training hyperedges over edges.}\label{tab:multitask_edges_vs_hyperedges}
        \begin{tabular}{llccc|ccc}
            \toprule
            & Type & \multicolumn{3}{c}{\shortstack{Train\\Unlabeled (iter 2)}} & \multicolumn{3}{c}{\shortstack{Train\\Unlabeled (iter 3)}}\\
            \cmidrule(lr){3-8}
            
            & & (1) & (2) & (3) & (1) & (2) & (3)\\
            \cmidrule(lr){3-8}
            \multirow{5}{*}{\rotatebox[origin=c]{90}{Edges}} & E: rgb & 42.85 & 5.04 & 10.37 & 32.79 & 21.66 & 12.40\\
             & E: hsv & 41.70 & 4.69 & 10.54 & 33.51 & 19.90 & 12.48 \\
             & E: softedges & 32.47 & 6.26 & 11.56 & 27.28 & 18.61 & 13.53 \\
             & E: softseg & 30.71 & 5.97 & 11.14 & 24.68 & 22.70 & 12.76 \\
             & E: ufo & 20.77 & 7.19 & 11.69 & 16.93 & 17.55 & 12.89 \\
             & DH-E: sseg & - & 6.25 & 11.39 & - & 19.00 & 12.93 \\
             & DH-E: depth & 29.24 & - & 12.22 & 24.11 & - & 13.79 \\
             & DH-E: norm & 30.56 & 6.17 & - & 26.35 & 21.15 & - \\
            \cmidrule(lr){2-8}
            & mean & 32.61 & 5.94 & 11.27 & 26.52 & \textbf{20.08} & 12.97 \\
            \midrule
            \multirow{5}{*}{\rotatebox[origin=c]{90}{Hyperedges}} & AH & 41.80 & 5.33 & 10.37 & 33.63 & 23.96 & 12.24 \\
            & AH-ufo & 41.96 & 5.16 & 10.78 & 33.82 & 21.10 & 12.72 \\
            & CH & 44.63 & 4.93 & 10.32 & 36.92 & 20.36 & 12.23 \\
            \\
            \cmidrule(lr){2-8}
            & mean & \textbf{42.80} & \textbf{5.14} & \textbf{10.49} & \textbf{34.79} & 21.81 & \textbf{12.40}\\
            \bottomrule 
            \vspace{-5mm}
        \end{tabular}
    \end{table}
    
\endgroup

\subsection{Impact of Different Ensemble Models}\label{sec:multipath_consensus}

We train all edges and hyperedges on the Train Unlabeled (iter 1) set and test different ensemble models for the task of semantic segmentation. Being the only task for which we have manually annotated ground truth, we focused on it for validating different ensembles. Our results are showcased in Table~\ref{tab:multipath_consensus_distilled}.

Both NGC~\cite{leordeanu2021semi} and CShift~\cite{haller2021cshift} models use only edges and relatively simple non-parametric ensemble models at nodes (NGC - simple average and CShift - non-parametric pixel-wise kernel weighted average). We bring a performance boost by 
adding hyperedges and also by allowing the ensembles to learn. As discussed in Sec. \ref{sec:learning_ensembles}, we propose one linear and 3 types of non-linear (neural nets) ensembles 
(Fig~\ref{fig:conv3ds-neural-aggregator}). Our experiments show that learning parametric ensemble models, even a simple linear one, improves significantly (above $2\%$ on average) over previously published work. Performance is reported on the Train Unlabeled (iter 3), which includes only the test scenes.

\begingroup
 \setlength{\tabcolsep}{2.7pt} 
    \begin{table}
        \centering
        \caption{Comparison to previous multi-task graph-based methods. We show considerable improvements by adding the proposed hyperedges (denoted with HE in the table) on top of existing work that uses only edges within their graph structure. We further report performance improvements by learning the proposed types of ensembles on top of both edges and hyperedges predictions (denoted by Ours). For this experiment, we considered solely the task of semantic segmentation. We report mean IoU (\% - higher values are better and bolded). The evaluation was done on each scene from the testing set and overall. LR stands for Logistic Regression (see Section~\ref{sec:learning_ensembles} for details).}
        \label{tab:multipath_consensus_distilled} 
        \begin{tabular}{lcccc}
            \toprule
            \multirow{2}{*}{Method} & \multicolumn{4}{c}{IoU($\uparrow$)}\\
            & Barsana & Comana & Norway & Mean \\
            \cmidrule(lr){1-5}
            NGC~\cite{leordeanu2021semi} (Mean) & 41.53 & 40.75 & 27.38 & 36.55 \\
            NGC (Mean) + HE & 42.61 & 42.17 & 27.96 & 37.58 \\  
            NGC~\cite{leordeanu2021semi} (Median)  & 39.25 & 37.41 & 27.01 & 34.56 \\
            NGC (Median) + HE & 44.34 & 38.99 & 22.63 & 35.32 \\
            \cmidrule(lr){1-5}
            CShift~\cite{haller2021cshift} (Mean) & 43.91 & 42.13 & 29.68 & 38.57 \\
            CShift (Mean) + HE & 44.71 & 43.88 & 30.09 & 39.56 \\
            CShift~\cite{haller2021cshift} (Median) & 43.30 & 40.62 & 29.51 & 37.81 \\
            CShift (Median) + HE & 46.27 & 43.67 & 29.09 & 39.68 \\
            \cmidrule(lr){1-5}
            LR (Ours) & 46.51 & \textbf{45.59} & \textbf{30.17} & \textbf{40.76} \\
            \cmidrule(lr){1-5}
            NN (v1) (Ours) & 45.53 & 42.92 & 28.37 & 38.94 \\
            NN (v2) (Ours) & 45.48 & 43.25 & 26.36 & 38.36 \\
            NN (v3) (Ours) & \textbf{48.21} & 44.85 & 28.94 & 40.67 \\
            \bottomrule
        \end{tabular}
    \end{table}
\endgroup


\subsection{Impact of Iterative Self-supervised Learning}\label{sec:iterative_learning}

To further highlight the benefits of learning through multiple iterations in a self-supervised manner, we test by progressively adding novel unlabeled data (learning iterations that are explained in Sec.~\ref{sec:dronescapes_dataset} and Sec.~\ref{subsec:semisup_hypergraph_learning}). 
After training each link in the hypergraph (Tab.~\ref{tab:multitask_edges_vs_hyperedges}) for each of the output tasks, we form multi-path consensus ensembles (Tab.~\ref{tab:multipath_consensus_distilled}), to produce pseudolabels for the second, self-supervised learning iteration, when we retrain all the edges and hyperedges on an expanded set of labels, which includes the supervised labels from Train Labeled, and the automatically generated labels from Train Unlabeled (iter 1) and Train Unlabeled (iter 2). We use again the newly retrained edges and hyperedges to form ensembles and pseudolabels for the additional Train Unlabeled (iter 3) set and continue retraining the single edges (\emph{rgb $\rightarrow$ task}) for a third self-supervised learning iteration. The results show considerable and consistent improvement at the level of the distilled edge \emph{rgb $\rightarrow$ task} in both accuracy and temporal consistency, on all 3 tasks (Tab.~\ref{tab:iterative_table}). We denote by $rgb-sup.$ the fully-supervised edge trained exclusively on the Train Labeled set.


\begingroup
    \setlength{\tabcolsep}{1pt} 
    \begin{table}
        \centering
        \caption{Iterative learning performance on the single task links. The evaluation was done on the test scenes. The reported performance is averaged.}\label{tab:iterative_table}
        \begin{tabular}{lcc|cc|cc}
            \toprule
            Type & \multicolumn{2}{c}{Semantic} & \multicolumn{2}{c}{Depth} & \multicolumn{2}{c}{Normals}\\
            & IoU ($\uparrow$) & Cons. ($\uparrow$) & L1 ($\downarrow$) & Cons. ($\uparrow$) & L1 ($\downarrow$) & Cons. ($\uparrow$)\\
            \cmidrule{2-7}
            rgb-sup. & 25.04 & 88.85 & - & - & - & -\\
            rgb-iter1 & 32.79 & 94.04 & 21.66 & 5.89 & 12.40 & 98.32\\
            rgb-iter2 & 37.26 & 95.72 & 17.34 & 7.06 & 11.93 & 98.87\\
            rgb-iter3 & \textbf{40.31} & \textbf{98.13} & \textbf{16.64} & \textbf{30.26} & \textbf{11.71} & \textbf{99.30}\\
            \bottomrule 
            
        \end{tabular}
    \end{table}
\endgroup

\begingroup
    \setlength{\tabcolsep}{2.3pt} 
    \begin{table*}[ht!]
        \centering
        \caption{Distilling a VisTransformer-based Expert's knowledge into the hypergraph. Experiments were done on novel scenes (test) for the task of semantic segmentation. Experiments show the advantages of using our hypergraph procedures in multiple learning phases and using an off-the-shelf method for the task of semantic segmentation. Performance evaluation was done at the level of the distilled direct edges. Bolded numbers are best.}\label{tab:sseg_sota_comparison}
        \begin{tabular}{lccccccccccc}
            \toprule
            Method & & & & \multicolumn{2}{c}{Barsana} & \multicolumn{2}{c}{Comana} & \multicolumn{2}{c}{Norway} & \multicolumn{2}{c}{Mean}\\
            \cmidrule(lr){5-12}
            & Phase 1 & Phase 2 & Phase 3 & IoU ($\uparrow$) & Cons. ($\uparrow$) & IoU ($\uparrow$) & Cons. ($\uparrow$) & IoU ($\uparrow$) & Cons. ($\uparrow$) & IoU ($\uparrow$) & Cons. ($\uparrow$)\\
            \midrule
            Mask2Former~\cite{cheng2022masked} & - & - & - & 56.77 & 95.48 & 59.84 & 97.26 & 41.71 & 98.23 & 52.77 & 96.99\\
            \midrule
            $rgb \rightarrow sseg$ (1) (Ours) & \cxmark & \checkmarkgreen & \checkmarkgreen & 57.86 & 98.46 & 57.82 & 98.88 & 42.95 & 99.51 & 52.88 & 98.95\\
            $rgb \rightarrow sseg$ (2) (Ours) & \checkmarkgreen & \checkmarkgreen & \cxmark & 58.02 & 98.22 & 60.14 & 98.70 & 42.55 & 99.42 & 53.57 & 98.78\\
            $rgb \rightarrow sseg$ (3) (Ours) & \checkmarkgreen & \checkmarkgreen & \checkmarkgreen & \textbf{59.16} & \textbf{98.58} & \textbf{60.49} & \textbf{99.06} & \textbf{43.14} & \textbf{99.55} & \textbf{54.26} & \textbf{99.06}\\
            \bottomrule 
            \vspace{-4mm}
        \end{tabular}
    \end{table*}
\endgroup

\begin{figure*}[ht!]
\centering
\includegraphics[scale=0.5]{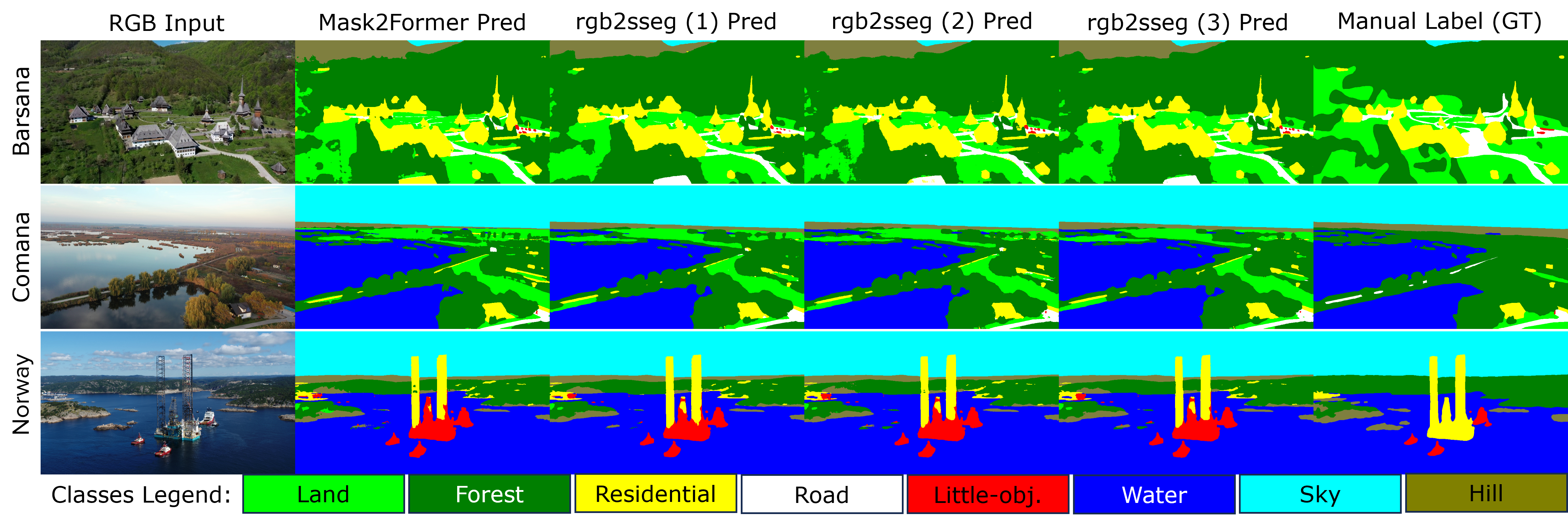}
\caption{Qualitative results on images from the testing set using Mask2Former labels as starting point for distillation. Even on a very dissimilar test scene w.r.t all other scenes (Norway), we obtain improved results over the baseline, and overall our method yields favorable numbers on all scenes, for both accuracy and temporal consistency, as shown in Tab. \ref{tab:sseg_sota_comparison}}.
\label{fig:qualitative_results}
\vspace{-8mm}
\end{figure*}

\subsection{Adapting to Novel Scenes}
\label{sec:sseg_sota_comparison}

We want to test the ability to improve over an initial state-of-the-art Expert Teacher and also adapt to novel test scenes. We focus on semantic segmentation and use the recent Mask2Former~\cite{cheng2022masked} as the SoTA Expert, pretrained on Mapillary Vistas~\cite{neuhold2017mapillary}. Being a similar scenario, we easily map the Mask2Former set of classes to ours. 


We consider three phases of learning in this setup, depending on the moment we use our self-supervised hypergraph learning. \textbf{Phase 1} denotes training the single $rgb \rightarrow seg$ edge, as presented so far, in two iterations of learning (before seeing any data from the test scenes). \textbf{Phase 2} refers to distilling $rgb \rightarrow seg$ edge (fine-tuning after Phase 1 or from scratch when Phase 1 is missing) on the output of Mask2Former on the test scenes. \textbf{Phase 3} refers to training $rgb \rightarrow seg$ during one iteration of self-supervised hypergraph learning, only on unlabeled data from test scenes, after Phase 2. In essence, Phase 1 and Phase 3 represent the same type of hypergraph learning with different starting points (initialization) for $rgb \rightarrow seg$, separated by Phase 2 of fine-tuning on Mask2Former output. Our goal is to evaluate the impact of the hypergraph model when applied at different stages, before and after Expert distillation. The results, presented in Tab.~\ref{tab:sseg_sota_comparison} and Fig.~\ref{fig:qualitative_results}, demonstrate the added value of the hypergraph in each case, with maximum effect when applied in both Phases 1 and 3. Interestingly, experiment 2 (hypergraph pretraining) is more effective than experiment 1 (hypergraph post-training), probably due to the additional data from the other scenes available in experiment 2. 
Also note that the added benefit of the hypergraph is significant at the level of temporal consistency (more results in the appendix), which suggests that the self-supervised consensus among multiple tasks strongly reduces classifier variance, potentially increasing reliability and trustworthiness. Moreover, the improvements are obtained by an edge that is two orders of magnitude smaller, in terms of the number of parameters, than the SoTA Mask2Former.

\vspace{-2mm}

\section{Conclusions}

\vspace{-2.5mm}

We introduced a novel multi-task self-supervised hypergraph model for learning in the case of very limited training data. Our theoretical contributions include the addition of hyperedges and parameterized ensemble learning, with proven experimental benefits. We also introduce Dronescapes, a large-scale video dataset for UAVs, which brings our experiments into the real world, different from the synthetic datasets used by prior works. With almost all annotations being automatically generated (except for a very small set of manually annotated frames for semantic segmentation) we show that by using hyperedges and learning ensembles of such hyperedges, we improve both accuracy and temporal consistency even though no temporal information is given. Our model is also effective when it uses as initial annotations the output of a state-of-the-art expert, as demonstrated by experiments on three different novel scenes, with no ground truth available for training.

\noindent \textbf{Acknowledgements:} This work was funded in part by UEFISCDI, under Projects EEA-RO-2018-0496 and PN-III-P4-ID-PCE-2020-2819, and by a Google Research Gift.

{\small
\bibliographystyle{ieee_fullname}
\bibliography{egbib}
}

\end{document}